\documentclass{bmvc2k}

\title{Towards Detection of Sheep Onboard a UAV}

\addauthor{Farah Sarwar$^{1,3,5}$}{farah.sarwar@aut.ac.nz}{0}
\addauthor{Anthony Griffin$^{2,3,5}$}{anthony.griffin@aut.ac.nz}{0}
\addauthor{Saeed Ur Rehman$^{1,3,5}$}{saeed.rehman@aut.ac.nz}{0}
\addauthor{Timotius Pasang$^{1,4,5}$}{timotius.pasang@aut.ac.nz}{0}
\addinstitution{
${^1}$Engineering Research Institute\\[1mm]
${^2}$High Perf. Comp. Research Lab.\\[1mm]
${^3}$Electrical and Electronic Eng. Dept.\\[1mm]
${^4}$Mechanical Eng. Dept.\\[1mm]
${^5}$School of Engineering, Computer and Mathematical Sciences\\[1mm]
Auckland University of Technology\\
Auckland, New Zealand}

\runninghead{Sarwar, Griffin, Rehman, Pasang}{Towards Det. of Sheep Onboard a UAV}

\def\etal{\emph{et al}\bmvaOneDot}

\begin{document}

\maketitle

\begin{abstract}
In this work we consider the task of detecting sheep onboard an unmanned aerial vehicle (UAV) flying at an altitude of 80~m.  At this height, the sheep are relatively small, only about 15 pixels across.  Although deep learning strategies have gained enormous popularity in the last decade and are now extensively used for object detection in many fields, state-of-the-art detectors perform poorly in the case of smaller objects.  We develop a novel dataset of UAV imagery of sheep and consider a variety of object detectors to determine which is the most suitable for our task in terms of both accuracy and speed.  Our findings indicate that a UNet detector using the weighted Hausdorff distance as a loss function during training is an excellent option for detection of sheep onboard a UAV.

\end{abstract}

\section{Introduction}
\label{sec:intro}
Wildlife or livestock monitoring can help zoologists and farmers to identify various animal-related issues. This perspective has motivated various researchers during the last few decades to analyse the videos recorded in various areas and provide required information to interested parties after processing the data. The approach of wildlife monitoring was started in 1991~\cite{jachmann1991evaluation} by Jachmann who estimated elephant densities using aerial samples. However, many challenges arose due to the diversity of background, species-specific characteristics, spatial clustering of animals~\cite{sileshi2008excess}. Various statistical and biological methodologies were used to handle these challenges. Some of the previously used techniques in wildlife life monitoring are: AdaBoost classifier~\cite{burghardt2006real}, power spectral-based techniques~\cite{parikh2013animal}, deformable part-based model (DPM), support vector machine (SVM)~\cite{van2014nature}, deep convolutional neural networks (CNN)~\cite{chamoso2014uavs,mckinlay2010integrating} and template matching algorithms~\cite{wichmann2010animal}. Vayssade \etal~\cite{vayssade2019automatic} has recently published work on goat activity detection using various classifiers with 74\% sensitivity of livestock detection.

Object detection in images is the accurate classification of the respective objects along with their location in the image. It can be performed using both machine learning and deep learning techniques.  In 2012, Ciresan demonstrated outstanding results for object classification on the NORB and CIFAR-10 datasets using neural networks~\cite{ciregan2012multi}. Similarly, Krizhevsky~\cite{krizhevsky2012imagenet} performed classification on ImageNet in 2012 using deep CNN (DCNN) and gave unmatched results. Since then extended versions of CNNs have been proposed such as, Region-based CNN (R-CNN)~\cite{girshick2014rich}, Fast R-CNN~\cite{girshick2015fast}, Faster R-CNN~\cite{ren2015faster}, you only look once (YOLO)~\cite{redmon2016you} single shot multibox detector (SSD)~\cite{liu2016ssd}, Mask R-CNN~\cite{he2017mask}, RetinaNet with focal loss~\cite{lin2017focal} and UNet with weighted Hausdorff distance (WHD)~\cite{ribera2018weighted}. Multiple versions of YOLO are available with unrivalled performance and are the fastest available algorithms for video processing but Redmon \etal{}~\cite{redmon2016you} have also mentioned that YOLO can give more localization errors as compared to Faster R-CNN. None of its versions is good enough for small object detection~\cite{redmon2016you}.

Small objects usually lack detailed features that contribute towards successfully training of CNNs. Researchers have used various machine learning algorithms for this task. Desabi \etal~\cite{desai2005small} used a multiple filter bank approach to detect and track small objects using motion as the main contributing factor. Meng \etal~\cite{meng2013adaptive} investigated infrared images for small object detection in the complex background using intensity, space and frequency domains of the images. Vard \etal~\cite{vard2012small} used object features from contour models and used them in an auto-correlation function to define an energy function. This energy function was then optimized for small object detection in full images. Recently researchers have started proposing new methodologies for small object detection using CNNs. Chen \etal~\cite{chen2016r} used the R-CNN to locate small objects in images relative to large objects. They augmented the state-of-the-art R-CNN algorithm with a context model and a small region proposal generator to improve the small object detection performance. They were able to improve the mean average precision by 29.8\%. 
Kong \etal~\cite{Kong_2016_CVPR} proposed HyperNet to fine-tune VGG16 and achieved 76.3\% mAP over Fast R-CNN. Their proposed method can perform object detection in 5~fps using GPUs and can be used for real-time applications. Faster R-CNN is explored by Eggert \etal~\cite{eggert2017closer} for company logo detection and they introduced an improved strategy to cover up the drawbacks of previous Faster R-CNN. Li \etal~\cite{li2017perceptual} proposed a perceptual generative adversarial network (Perceptual GAN) that narrows representation differences of small objects from the large ones. Yancheng Bai \etal~\cite{Bai_2018_ECCV} proposed an end-to-end multi-task generative adversarial network (MTGAN), which up-samples the small images to recover necessary fine details.

In this work, we tackle the problem of detecting and counting sheep from unmanned aerial vehicle (UAV) imagery.  As mentioned above, this is a challenging problems as the UAV flies at an altitude of 80~m and the sheep are thus whitish  blobs of about 15 pixels across.  Our previous work in this area~\cite{sarwar2018detecting} did not show promising results for the CNN object detectors considered.  Here we consider a much wider range of CNN object detectors, to investigate the true state-of-the-art performance.  Our final goal is for the object detection to be performed onboard the UAV, so we also consider the detectors' suitability for an embedded implementation.

\section{Methods}
There are two main categories of CNN-based object detectors: two-stage and one-stage detectors.  Two-stage detectors usually generate hundreds of region proposals per training image and train the network using these region proposals instead of using the actual image directly.  The state-of-the-art two-stage object detector is still R-CNN which trains the CNN with 2000 region proposals. Training and testing time using R-CNN is much slower---when compared to Fast R-CNN and Faster R-CNN---but it performs more accurately, generating fewer false positive detections.
One-stage detectors, like YOLO and SSD, are faster but they mostly work at the cost of accuracy. The main strategy of these detectors is the dense sampling of object locations using different scales and aspect ratios, and then dealing the task of object detection as a single regression problem. Taking image pixels as input, usually bounding box coordinates and respective confidence values are generated as output~\cite{redmon2017yolo9000}. 

\subsection{Two-Stage Detectors}
\label{sec:twoStage}
The current state-of-the-art two-stage object detector is R-CNN.  We have used a modified version of R-CNN for training AlexNet, GoogLeNet and two newly-proposed networks. Instead of using 2000 region proposals in the initial stage, 1000 region proposals are used with positive and negative region overlap of (0.5,1) and (0.1,0.2) respectively. The actual regions of interest (ROI) are very small in the training images and it is expected that the more positive overlapped region proposals the better it will perform in object detection. The classification task is performed using softmax and cross entropy instead of an SVM to avoid the time to train the SVM classifier. The softmax outputs the result in terms of class probabilities and the cross entropy loss function uses output probabilities to compute the loss with respect to the actual class labels.

\begin{figure}
\begin{center}
\includegraphics[width=0.8\textwidth]{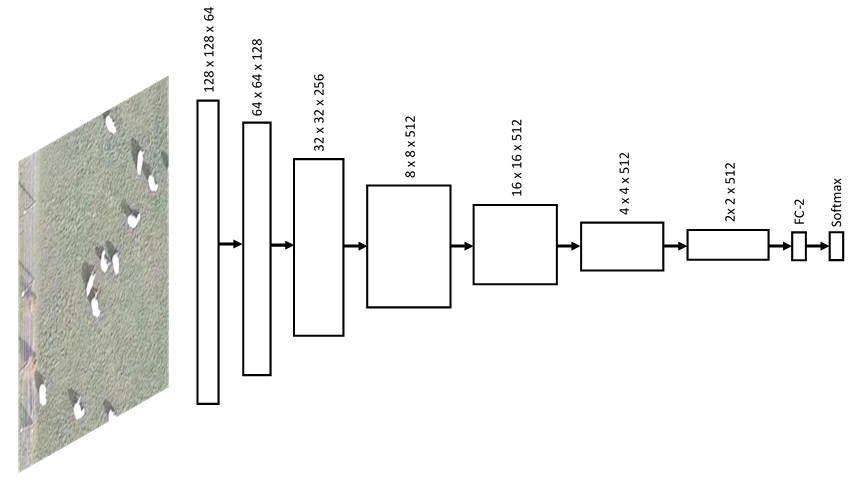}
\end{center}
   \caption{A 7-layer proposed network, which is trained using R-CNN.}
\label{fig:MyNet}
\end{figure}

In addition to AlexNet and GoogLeNet, we have proposed two very different network architectures to observe the impact of network architecture on performance metrics. One of the networks, Network-I, has only one convolutional (CONV) layer with $11\times 11 \times 96$ kernels, one fully connected (FC) layer and a softmax layer. This is the simplest architecture of CNN that can be used. 
The purpose of using only one CONV layer is to verify if smaller networks can work more efficiently for small object detection instead of larger and complex networks. Performance of different combinations of CONV and FC layers, kernel sizes and network topology has also been compared in terms of precision and recall in our previous work~\cite{sarwar2018detecting}. It was observed that using a five CONV- and two FC-layer network does not improve the results in a noticeable manner. However, those networks were trained on very small data sets of roughly 400 images. Instead of training all of those networks on bigger training data, a complex network is proposed as Network-II. It is a seven-layer network and is shown in Figure~\ref{fig:MyNet}.
The number of kernels and features are shown above each layer. Each layer has a block of $3\times 3$ convolution with stride 1, batch normalization, ReLU and maxpool layer of dimension $2\times 2$ and stride 2. The input image is $256 \times 256 \times 3$ and the  dimension of the features keep on decreasing by a factor of 2 until the last CONV layer. Only the $2 \times 2 \times 512$ feature matrix is linked with 2 neurons of the FC layer. Then the softmax layer performs classification on the output retrieved from the FC layer. It is designed in a way that features keep on getting denser towards the end of the network to keep only the objects' contour information. This network is proposed for those small objects which mostly exhibit only contour information and lacks highly detailed information within them. It has not been tested on other larger or smaller objects yet, but it is expected that it can be used for higher order of classes. The dimension of the FC layer needs to be adjusted according the number of object classes. We have two objects in our dataset---sheep and background---so the dimension of the FC layer is set to two. This network is more flexible than the network in the next section, in that it can be used for any number of classes, and training images of any dimension can be used. However, for comparing the performance metrics, size of training images, network settings and training parameters are kept same for all. 

\subsection{One-Stage Detector}
\label{sec:oneStage}
\begin{figure}
\begin{center}
\includegraphics[width=0.8\textwidth]{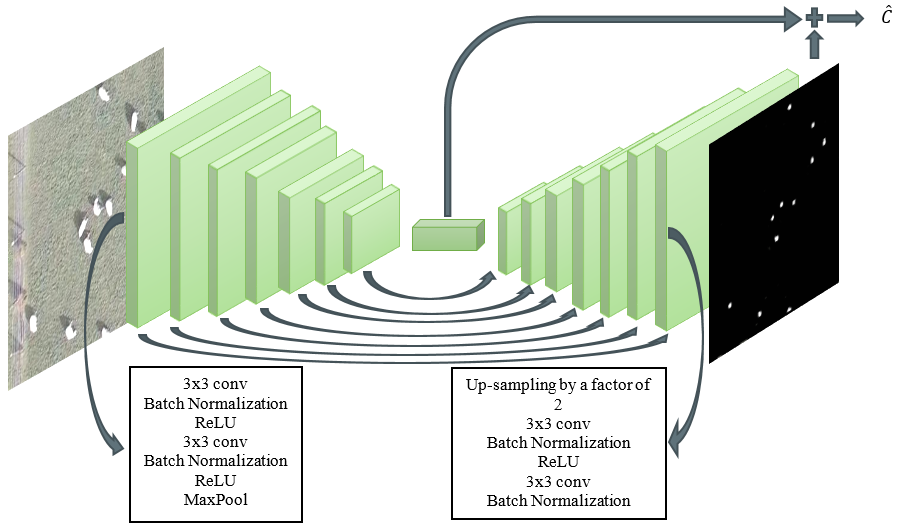}
\end{center}
   \caption{The complete UNet architecture for livestock detection. The blocks added in the lower part show the full configuration of each kernel in both sections.}
\label{fig:UNet}
\end{figure}

As discussed above, existing one-stage detectors like SSD and YOLO can be used for real-time applications but give many false positive detections on background~\cite{redmon2016you} and cannot be used in the areas where accuracy cannot be compromised. In recent work, Ribera \etal~\cite{ribera2018weighted} used a fully connected network---also know as UNet---and trained it as a one-stage network using the weighted Hausdorff distance (WHD) and smooth L1 loss as the loss function. Instead of using bounding boxes, it uses centroids of the objects as ground truth labelling and gives a probability map of training image along with a regression term. The UNet model they have used is similar to the one shown in Figure~\ref{fig:UNet}. The upper part of the Figure shows the full architecture of the U-Net model. While the lower part shows the configuration of a single kernel for respective parts. There are 64, 128, 256, 512, 512, 512, 512 and 512 kernels in the contraction part and correspondingly, 512, 512, 512, 512, 256 and 128 kernels in the expansion part.  The features from each convolutional block are concatenated with the features in the expansion part. The linear layer at the deepest level is used to regress the object estimation. The final layer is a $1\times 1$ CONV layer for mapping the 128-component feature maps from the second to last layer as an output probability map. The number of objects is estimated by passing the concatenated  $1\times 1\times 512$ feature vector and $256\times 256\times 1$ probability map through the $1\times 1\times 1$ fully connected layer. The output of this layer was then rectified using the SoftPlus non-linearity function, as shown in (\ref{c}). It is then rounded to nearest number to get the estimated number of detected objects.
\begin{equation} \label{c}
\hat{C} = \log(1 + e^{s})
\end{equation}
This methodology not only locates small objects successfully but can also be used for various complexly shaped objects. The only limitation of this technique is that it can be used for only one object class and needs some modification for multi-class classification and detection. Unlike all other object detection algorithms, it uses centroid-based ground truth data. This property makes it more convenient to use it in the cases where objects are complexly-shaped, small, and overlapping each other. Centroid based datasets are required for training this network. The WHD which is proposed in~\cite{ribera2018weighted} is as follows:
\begin{equation} \label{whd}
L_{T}(p,Y) = \frac{1}{|{X}|+\epsilon}
\sum_{x\,\in\,\Omega} p_x\min_{y\,\in\, Y} d(x,y) + \frac{1}{|Y|}
\sum_{y\,\in\,Y}\min_{x\,\in\,\Omega} \frac{d(x,y)+\epsilon} {p_x^\alpha+\frac{\epsilon}{d_\textrm{max}}} + L_{1}(C-\hat{C})
\end{equation}
Where 

\({X}\) = Estimated object locations as coordinates

\(Y\)= Actual object locations 

\(d(x,y)\) = Euclidean distance between estimated and actual locations in image

\(p_x\) = Probability of estimated object at location \(x\)

\(\epsilon\) = Correction factor to avoid infinity loss in case of no estimated object

\(d_\textrm{max}\) = Maximum distance between extreme points of \(X\) and \(Y\)

\(L_{1}\) = Smooth L1 loss function as shown in (\ref{L1})

\begin{equation}  \label{L1}
L_{1} = \left \{
  \begin{tabular}{cc}
  \(0.5x^2\), & for \(|x| < 1\)\\
  \(|x| -  0.5\), & for \(|x| \geq 1\)\\ 
  \end{tabular}
  \right\} 
\end{equation}

The only difference in training between the networks in Section~\ref{sec:twoStage} and this network is the use of bounding boxes as ground truth for the two-stage networks and centroids for UNet.  

\section{Dataset and Training}
\label{sec:dataset}
To our knowledge there are no existing publicly-available datasets of imagery of sheep on farms taken by a drone, so we had to create our own.  We used a DJI Phantom 3 Pro which we flew over paddocks of sheep with its camera pointed straight down.  Although the legal altitude limit for UAVs in New Zealand---and many other countries---is 120~m, for this work we chose 80~m as it offered a good compromise between enough detail on the sheep and enough of the paddock to be useful, as well as incurring minimal disturbance to the sheep.  Nonetheless, at this height in a $2048\times 1080$ frame of video, an adult sheep is a rather nondescript whitish blob about 20 pixels long and 10 pixels wide.  Videos were recorded at different times of day, and in different weather conditions, in order to capture the diversity in illumination, shadow size and background colour.  After data augmentation techniques were used, a full dataset of 22,742 images of size  $256\times256\times3$ were obtained, and labelled with bounding boxes and centroids.  This dataset was then divided into 80\%, 10\% and 10\% for training, validation, and testing, respectively.

All of the networks were trained using stochastic gradient descent method with learning rate $1\times10^{-4}$ and momentum 0.9. The networks were trained for 500 epochs while using batch size of 10 images. The smaller batch size was due to the limitation of available memory for network training. It was validated after every two epochs over full the validation set. 
All training and data inference sessions were performed on an NVIDIA GeForce GTX 1080, which has 8 GB of memory, 2560 CUDA cores and performs 16-bit, 128 giga floating point operations per second (GFLOPS).

As discussed in Sections~\ref{sec:twoStage} and \ref{sec:oneStage}, we evaluated the performance of 5 different networks: Network-I, Network-II, AlexNet, GoogLeNet and UNet.  Networks I and II as well as UNet, were trained from scratch. AlexNet and GoogLeNet were previously trained on ImageNet and were already able to classify 1000 object categories. After modifying the last two layers to suit our application, both were then retrained on our dataset.  

\section{Performance Evaluation}
\label{sec:performance}

In evaluating the performance of these networks in detecting sheep, we focused on the precision, recall and $F_1$-score, defined as follows:
\begin{equation}
    \textrm{precision} = \frac{\textrm{TP}}{\textrm{TP} + \textrm{FP}} 
    \qquad\quad
    \textrm{recall} = \frac{\textrm{TP}}{\textrm{TP} + \textrm{FN}} 
    \qquad\quad
    F_1 = 2\cdot\frac{\textrm{precision}\cdot\textrm{recall}}{\textrm{precision}+\textrm{recall}} 
\end{equation}
where TP, FP and FN are the numbers of true positives, false positives and false negatives, respectively.

Precision can be thought of as "the number of detected objects that are actually sheep", and recall as "the number of sheep that are actually detected".  The $F_1$-score is their harmonic mean, and a way to consider both the precision and recall in one number.  Table \ref{table:metrics} presents these metrics for the 5 networks considered, along with the root mean square error (RMSE) of the locations, and the time per image (TPI) for the inference.

\begin{table}[h]
\vspace*{0mm}
\begin{center}
\begin{tabular}{|c|c|c|c|c|c|}
 \hline
 Networks & Precision & Recall & $F_1$-score & RMSE & TPI\\
 \hline
 UNet & \textbf{96.20\%} & \textbf{90.14\%} & \textbf{93.07\%} & \textbf{0.783} & \textbf{0.14 s} \\
 Network-I & 88.77\% & 80.45\% & 84.38\% & 1.961 & 0.20 s\\
 Network-II & 91.21\% & 81.91\% & 86.31\% & 1.819 & 2.22 s\\
 GoogLeNet & 80.79\% & 81.67\% & 81.23\% & 1.899 & 3.53 s \\
 AlexNet & 78.74\% & 81.37\% & 80.03\% & 2.132 & 1.31 s\\
 \hline
\end{tabular}
\end{center}
\caption{Performance metrics of the five networks}
\label{table:metrics}
\end{table}

The UNet clearly outperforms the other networks, in all metrics making it the obvious choice for our task.  Network-II is the second best performing in terms of accuracy, but is considerably slower than UNet.
\begin{figure}
\begin{center}
\includegraphics[width=2.5cm, height=3cm]{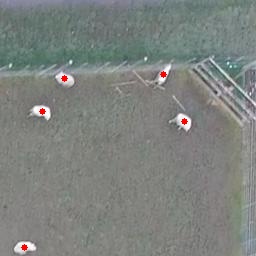}
\includegraphics[width=2.5cm, height=3cm]{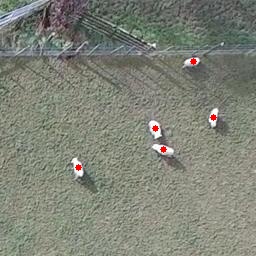}
\includegraphics[width=2.5cm, height=3cm]{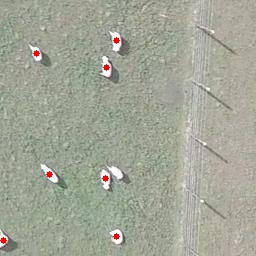}
\includegraphics[width=2.5cm, height=3cm]{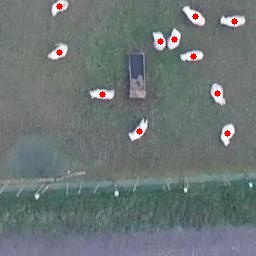}
\includegraphics[width=2.5cm, height=3cm]{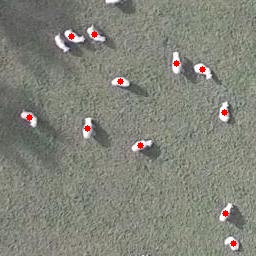}
\includegraphics[width=2.5cm, height=3cm]{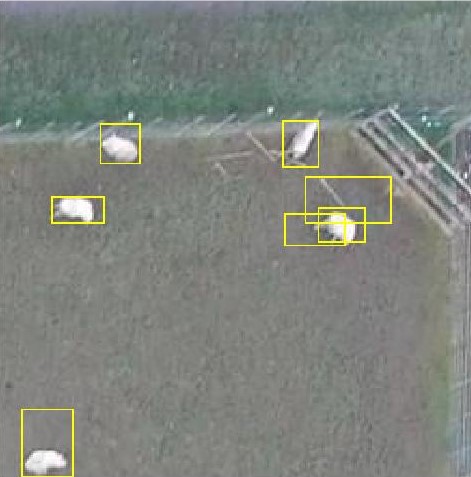}
\includegraphics[width=2.5cm, height=3cm]{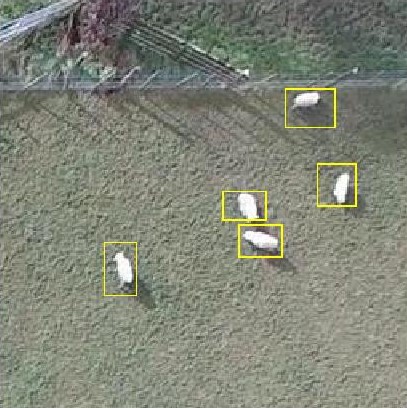}
\includegraphics[width=2.5cm, height=3cm]{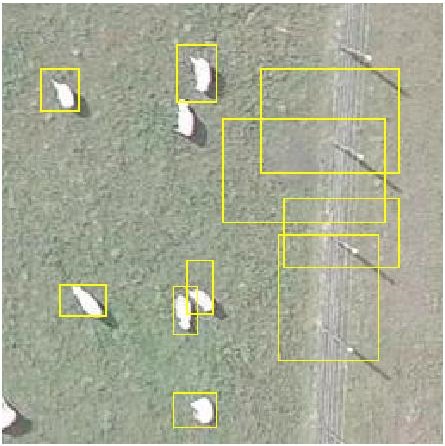}
\includegraphics[width=2.5cm, height=3cm]{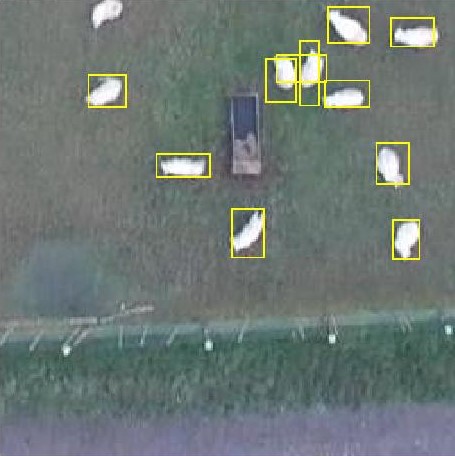}
\includegraphics[width=2.5cm, height=3cm]{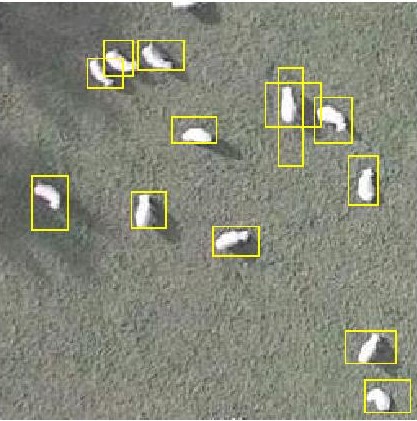}
\includegraphics[width=2.5cm, height=3cm]{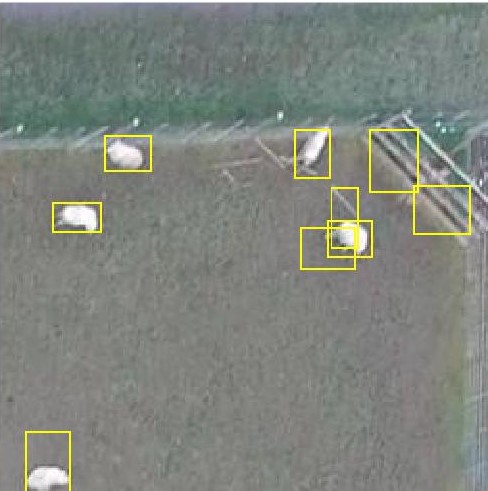}
\includegraphics[width=2.5cm, height=3cm]{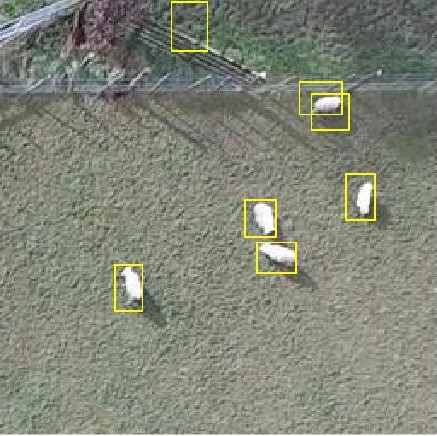}
\includegraphics[width=2.5cm, height=3cm]{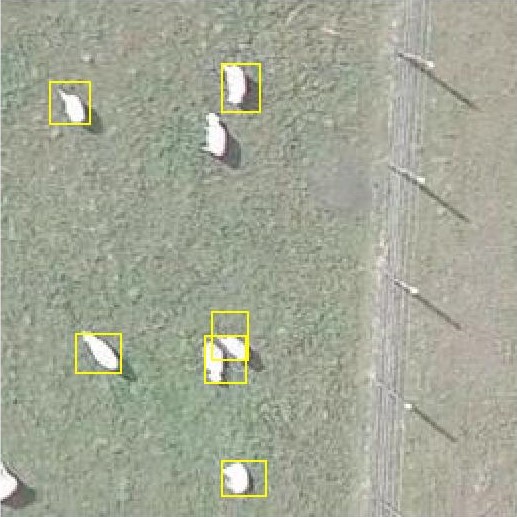}
\includegraphics[width=2.5cm, height=3cm]{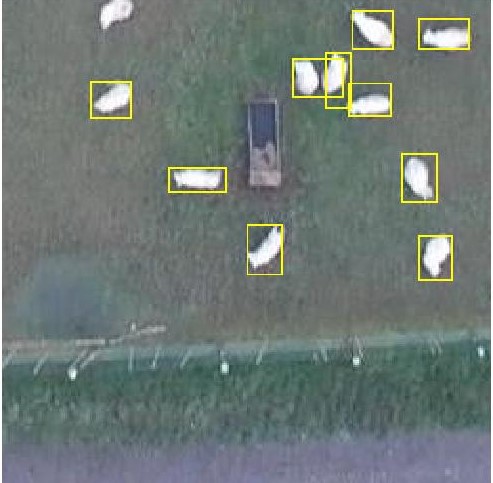}
\includegraphics[width=2.5cm, height=3cm]{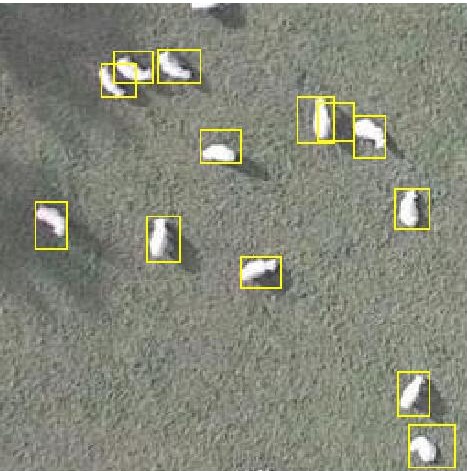}
\includegraphics[width=2.5cm, height=3cm]{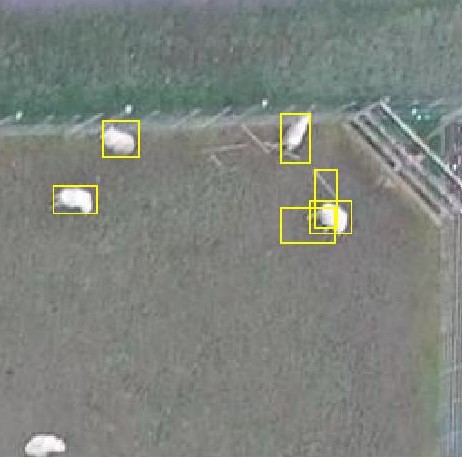}
\includegraphics[width=2.5cm, height=3cm]{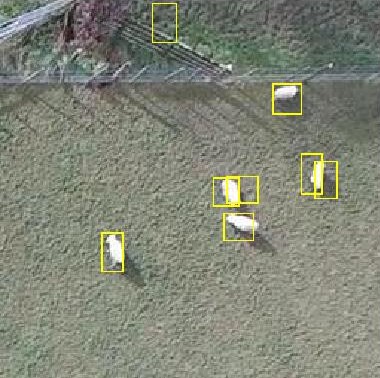}
\includegraphics[width=2.5cm, height=3cm]{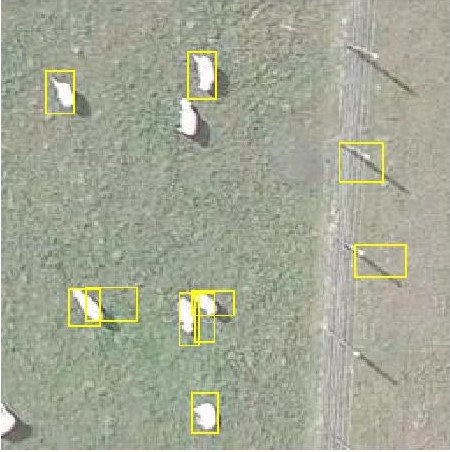}
\includegraphics[width=2.5cm, height=3cm]{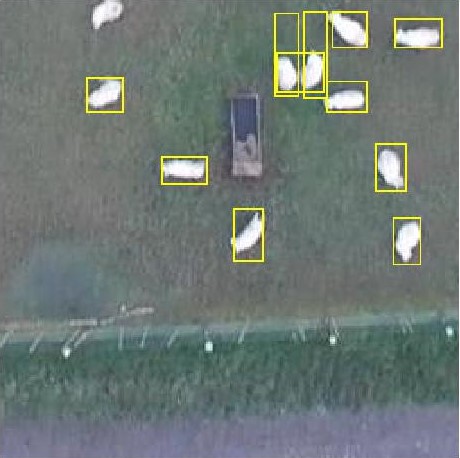}
\includegraphics[width=2.5cm, height=3cm]{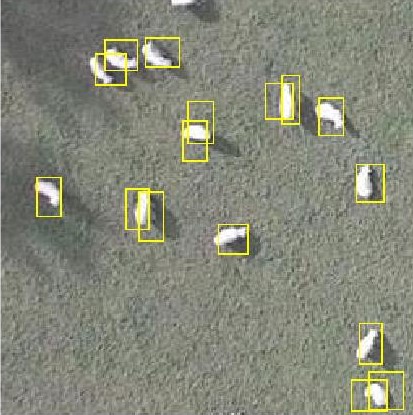}
\includegraphics[width=2.5cm, height=3cm]{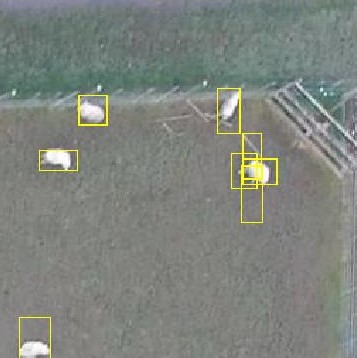}
\includegraphics[width=2.5cm, height=3cm]{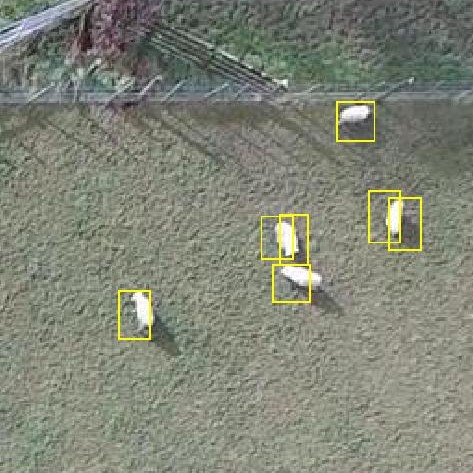}
\includegraphics[width=2.5cm, height=3cm]{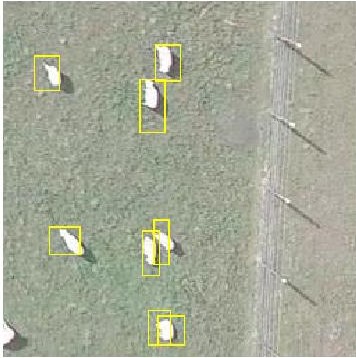}
\includegraphics[width=2.5cm, height=3cm]{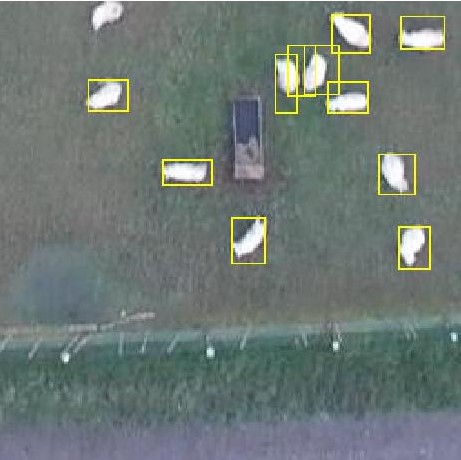}
\includegraphics[width=2.5cm, height=3cm]{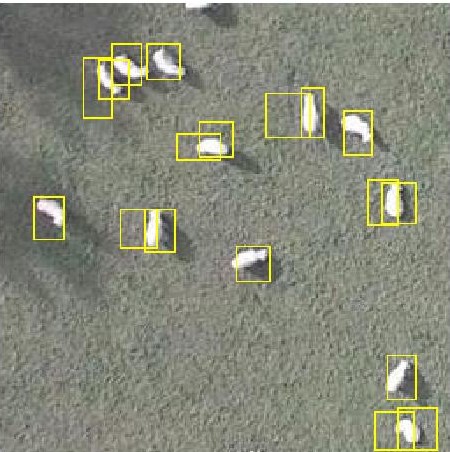}
\caption{Example output test images from UNet, Network-I, Network-II, AlexNet and GoogLeNet in progressive rows respectively.  Note that UNet provides estimated centroids, while the other four networks provide estimated bounding boxes.}
\end{center}
\label{fig:results}
\end{figure}
Figure \ref{fig:results} shows the output of some example test images. Networks I and II have more false positive detections on fences, whereas AlexNet and GoogLeNet exhibit multiple object detections at the same points and less false positive errors.  UNet  has  the  least  detection  errors, and only fails when sheep are very close together.  

We are confident that sheep detection and counting could be performed onboard a custom UAV in real- or near real-time, using the UNet detector and hardware such as the NVIDIA Jetson Nano.  The Jetson Nano  provides 472 GFLOPS and is specially designed for the implementation of AI-based applications like detection, classification and segmentation. It has 128 CUDA cores, 4 GB 64-bit RAM and 16 GB Flash storage, and uses only 5~W of power.

\section{Conclusions}
\label{sec:conclusions}
In this paper, we have investigated a variety of object detectors to find which is the most suitable to detect sheep onboard a UAV.  A detector based on UNet using the weighted Hausdorff distance as a loss function during training was shown to be an excellent option, and should be able provide real- or near real-time sheep detection onboard a UAV.

\bibliography{bmvcbib}
\end{document}